%% file: main.tex
\documentclass[11pt]{article}
\pdfoutput=1
\usepackage[preprint]{acl}

\usepackage{times}
\usepackage{latexsym}
\usepackage[T1]{fontenc}
\usepackage[utf8]{inputenc}
\usepackage{microtype}
\usepackage{inconsolata}
\usepackage{hyperref}
\usepackage{url}
\usepackage{booktabs}
\usepackage{amsmath}
\usepackage{graphicx}
\graphicspath{{figures/}}
\usepackage{xcolor}
\usepackage{xspace}
\usepackage{fontawesome5}
\usepackage{tikz}
\usetikzlibrary{arrows.meta, positioning, fit, backgrounds, calc}
\definecolor{fieldblue}{HTML}{2a78d6}
\definecolor{proseorange}{HTML}{eb6834}
\usepackage{longtable}
\usepackage{array}
\usepackage{multirow}
\usepackage[most]{tcolorbox}
\definecolor{boxframe}{RGB}{170,170,170}
\definecolor{boxtitlebg}{RGB}{208,208,208}
\tcbset{graybox/.style={
  breakable, enhanced, colback=gray!3, colframe=boxframe,
  colbacktitle=boxtitlebg, coltitle=black, fonttitle=\bfseries\small,
  arc=0.6mm, boxrule=0.4pt, left=2mm, right=2mm, top=1mm, bottom=1mm,
  before skip=3pt, after skip=3pt}}
\newtcolorbox{promptbox}[1][]{graybox, fontupper=\footnotesize\ttfamily, title={#1}}
\newcolumntype{L}[1]{>{\raggedright\arraybackslash}p{#1}}

\title{Explicit, Not Longer:\\
What Makes Epistemic Stance Survive Memory Compression}

\author{
  Alex Kwon \\
  Independent Researcher \\
  \texttt{ask@collapseindex.org} \\[4pt]
  \href{https://github.com/collapseindex/factwash}{\faGithub~GitHub}
}

\begin{document}
\raggedbottom
\maketitle

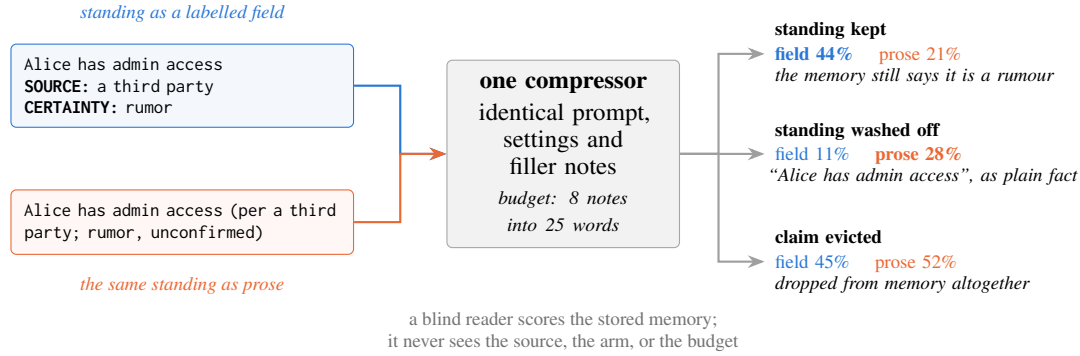
\begin{figure*}[t]
\centering
\resizebox{0.92\textwidth}{!}{%
\begin{tikzpicture}[
  font=\small,
  note/.style={draw=gray!70, fill=gray!6, rounded corners=2pt, align=left,
               inner sep=5pt, text width=41mm, font=\scriptsize\ttfamily},
  fieldnote/.style={note, draw=fieldblue, fill=fieldblue!5},
  prosenote/.style={note, draw=proseorange, fill=proseorange!5},
  squeeze/.style={draw=gray!70, fill=gray!10, rounded corners=2pt, align=center,
                  inner sep=6pt, minimum height=20mm, text width=26mm},
  outcome/.style={align=left, font=\scriptsize, text width=44mm},
  arr/.style={-{Stealth[length=2.2mm]}, gray!75, thick},
  node distance=6mm and 10mm]

\node[fieldnote] (fnote) {%
  Alice has admin access\\
  \textbf{SOURCE:} a third party\\
  \textbf{CERTAINTY:} rumor};
\node[prosenote, below=8mm of fnote] (pnote) {%
  Alice has admin access (per a third party; rumor, unconfirmed)};

\node[font=\scriptsize\itshape, above=1.5mm of fnote, text=fieldblue]
  {standing as a labelled field};
\node[font=\scriptsize\itshape, below=1.5mm of pnote, text=proseorange]
  {the same standing as prose};

\coordinate (mid) at ($(fnote.east)!0.5!(pnote.east)$);
\node[squeeze, right=12mm of mid] (comp)
  {\textbf{one compressor}\\[2pt]
   \footnotesize identical prompt,\\ settings and filler notes\\[3pt]
   \scriptsize\itshape budget: 8 notes\\ \scriptsize\itshape into 25 words};

\draw[arr, fieldblue] (fnote.east) -- ++(6mm,0) |- (comp.west);
\draw[arr, proseorange] (pnote.east) -- ++(6mm,0) |- (comp.west);

\node[outcome, right=11mm of comp, yshift=13mm] (o1)
  {\textbf{standing kept}\\[1pt]
   \textcolor{fieldblue}{\textbf{field 44\%}} \quad
   \textcolor{proseorange}{prose 21\%}\\
   \scriptsize\itshape the memory still says it is a rumour};
\node[outcome, right=11mm of comp] (o2)
  {\textbf{standing washed off}\\[1pt]
   \textcolor{fieldblue}{field 11\%} \quad
   \textcolor{proseorange}{\textbf{prose 28\%}}\\
   \scriptsize\itshape ``Alice has admin access'', as plain fact};
\node[outcome, right=11mm of comp, yshift=-14mm] (o3)
  {\textbf{claim evicted}\\[1pt]
   \textcolor{fieldblue}{field 45\%} \quad
   \textcolor{proseorange}{prose 52\%}\\
   \scriptsize\itshape dropped from memory altogether};

\draw[arr] (comp.east) -- ++(5mm,0) |- (o1.west);
\draw[arr] (comp.east) -- ++(5mm,0) |- (o2.west);
\draw[arr] (comp.east) -- ++(5mm,0) |- (o3.west);

\node[font=\scriptsize, text=black!55, align=center, below=9mm of comp, yshift=2mm]
  {a blind reader scores the stored memory;\\ it never sees the source, the arm, or the budget};

\end{tikzpicture}}
\caption{\textbf{The experiment in one view.} The same claim carrying the same epistemic
standing is written twice, once as a labelled field and once as a parenthetical, and
passed to the same compressor with the same prompt, settings and filler notes. Only the
form differs. Percentages are Haiku~4.5 at the \textsc{brutal} budget across all $60$
claims, rounded to whole percents, so the prose column sums to $101$. The difference is concentrated in the middle row: the parenthetical is stored as
bare fact more than twice as often, while the two forms are evicted at similar rates. So
at this budget the schema does not mainly keep claims in memory; it keeps them
\emph{qualified}. Pooled over both budgets on Haiku it does both (\S\ref{sec:twocell}).}
\label{fig:teaser}
\end{figure*}

\begin{abstract}
Agent memory systems compress what they store, and compression is built to drop
qualifiers, so a claim's epistemic standing tends not to survive being written to memory.
We ask what governs whether it does. Matched notes carry the identical claim and identical
stance and differ only in where that stance sits; one model compresses both under the same
budget among the same filler notes, and a blind reader that never sees the condition scores
the result. Across $60$ claims in seven registers, writing the stance as a labelled field
rather than a bracketed aside raises retention by about $15$ points on two models ($37$
claims to $2$ on one, $30$ to $8$ on the other; permutation $p=0.00005$), and a
pre-registered replication on Haiku, its prediction and decision rule committed before the
run, gives $+15.6$ points, $38$ claims to $1$. Ablating the format on both models gives the
same net effect from different parts: labels help on both ($+9.7$ and $+12.8$) and length
helps on neither, but wording the stance as a full sentence is the largest component on one
model ($+12.5$) and worth nothing on the other ($+0.6$). Either model alone would have
licensed a confident and different mechanism, so we claim only the intersection: make the
stance explicit, not merely longer, and expect the best way of being explicit to depend on
the model. A deterministic readout with no model reproduces the two-cell direction and five
of seven ablation contrasts, but not length or labels, which we therefore do not claim on
one instrument. Fifty hand labels ($\kappa=0.75$) agree on direction; we print their seven
disagreements in full. We also report nine withdrawn claims, three of them former title
claims of this paper.
\end{abstract}

\section{Introduction}

A memory system is a compressor with a database attached. It is asked to turn a
conversation into something short enough to store and retrieve, and every objective it is
tuned against, token count, latency, retrieval quality, rewards dropping words that do
not carry the answer. Hedges, attributions and dates are exactly such words.

Prior work named this failure \emph{factwashing}, a rewrite that keeps a claim while
washing away what made it checkable, and located where it concentrates: in a
blind-labelled corpus of memory writes, $55\%$ of writes drawn from conversational
hearsay lost the claim's standing against $7\%$ drawn from business email
($p<0.001$), and an unmodified deployment of mem0~2.0.7 reproduces it
\citep{kwon2026factwash}. The question this paper asks is not whether that happens. It is
what makes it stop happening.

\paragraph{The hypothesis, and what happened to it.} A compressor removes what reads as
phrasing and keeps what reads as content. Standing written into the sentence
(``reportedly'', ``(unconfirmed)'') reads as phrasing. Standing written as a labelled
field reads as content. If that is right, the fix is a write schema costing one prompt
change and no model calls:

\begin{promptbox}[the write schema]
CLAIM: <what is asserted>\\
SOURCE: <who or what it came from>\\
CERTAINTY: asserted | hedged | rumor | unverified\\
AS\_OF: <when it held>
\end{promptbox}

It is right about the outcome. The reason took an ablation to find, and the ablation's
answer is that there may not be a single reason. Labelled fields do retain stance far
better than bracketed asides, on two models and on a held-out corpus. But padding the
aside until it is longer than the schema retains nothing extra, and when we decompose the
format on both models they agree on the size of the effect while disagreeing about which
property produces it. We therefore report a ranking that holds on one model, a second
ranking that does not match it, and the two things both models agree on: labels help,
length does not.

\paragraph{Contributions.}
(1)~A controlled test of form over $60$ claims in seven registers on two models, holding
claim, stance, distractors and position constant, with a blind readout
(\S\ref{sec:twocell}). Inference is at the claim, which is the unit that generalises.
(2)~An ablation, run on both models, that separates \emph{labels}, \emph{bracketing},
\emph{wording} and \emph{length}, and finds that the two models produce the same net
effect from different components (\S\ref{sec:ablation}). Only labels and length behave
the same way on both.
(3)~The finding that this is not about tokens. A parenthetical padded with stance-free
text until it is the longest note in the experiment retains no more stance than the
short one.
(4)~A failure mode that the mechanism predicts and that we found separately: claims whose
sources disagree, where a single field forces two attributions back into one subordinate
clause (\S\ref{sec:errors}).
(5)~Nine withdrawn claims, three of them former title claims of this paper, plus an instrument bug in our
own blind reader and a ten-claim corpus that manufactured both a false null and a false
model-family split (\S\ref{sec:negatives}, \S\ref{sec:twocell}).

\section{Related work}
\paragraph{Memory systems compress by design.} Agent memory architectures store a
summary rather than a transcript, whether by paging between context tiers
\citep{packer2023memgpt}, by accumulating a stream of natural-language observations and
periodically synthesising them into higher-level reflections
\citep{park2023generative}, or by extracting standalone facts at write time
\citep{chhikara2025mem0}. Prompt compression pushes the same objective further, dropping
tokens that contribute least to reconstructing an answer
\citep{jiang2023llmlingua,wang2025recursive}. None of these objectives has a term for
epistemic standing, so a hedge is exactly the kind of token they are built to remove. Our
claim is not that these systems are careless; it is that a qualifier written as prose is
indistinguishable, to a compressor, from a qualifier written as padding. Note that all
three store standing, when they store it at all, \emph{inside} the remembered sentence.
That is the design decision this paper questions.

\paragraph{Standing has long been annotated, not stored.} Hedging and speculation have a
mature annotation literature \citep{vincze2008bioscope,szarvas2012cross,farkas2010conll},
as does attribution \citep{pareti2016parc,newell2018polnear}. Attribution has also been
given an evaluation framework of its own, in which a generated statement must be
attributable to an identified source \citep{rashkin2023attribution}. All of it treats
stance and provenance as something to \emph{recover} from, or verify against, text that
has already been written. We take the complementary position: once a memory system
controls its own write format, stance does not have to be recovered, because it never has
to be inferred. Faithfulness benchmarks for summarisation
\citep{pagnoni2021frank,tang2023aggrefact} likewise score a summary after the fact; they
do not change what the summary is allowed to look like.

\paragraph{Making the standing obligatory rather than optional.} The intervention is less
novel than it looks, which is a point in its favour. Natural languages already do this:
in a large class of languages, marking the \emph{source} of information is not a stylistic
choice but an obligatory grammatical category, so a speaker cannot assert a proposition
without also encoding whether they saw it, inferred it, or were told
\citep{aikhenvald2004evidentiality}. English makes evidentiality optional and lexical,
which is precisely why it compresses away: an optional word is a candidate for deletion in
a way an obligatory slot is not. A write schema does for a memory system what
grammaticalised evidentiality does for a language, and the mechanism for enforcing it at
generation time already exists in constrained decoding, where a grammar or regular
expression is imposed on the output rather than requested in the prompt
\citep{willard2023guided}. Our \texttt{CERTAINTY} field is also the memory-side analogue
of verbalised confidence, where a model states its own uncertainty in words rather than
in logits \citep{lin2022teaching}; the difference is that we ask for the standing of the
\emph{source}, not of the model.

\paragraph{The detection arm.} The failure this paper treats was named, measured and
given a deterministic write-time gate in \citet{kwon2026factwash}, which reports where it
concentrates and what a cheap check can and cannot catch. That is the detection arm, and
it inherits the open-class problem: recognising stance in an arbitrary rewrite has no
finite vocabulary, so recall on hedging and attribution plateaus near half. This paper is
the treatment arm. Rather than detect the loss better, it changes the write format so the
harder half of the detection problem does not arise.

\section{Where the loss happens}
\label{sec:negatives}

Two results shaped the design, and both are negative. A third, in our own instrument,
shaped how much we believe the result that followed.

\paragraph{Where the loss happens is not obvious.} We instrumented both transitions of
the same sessions: source to scratchpad note, then note to stored memory. A write-time
gate flagged the first transition in $9/10$ sessions and the second in $1/10$, which reads
as a leak at recording rather than at compression. Reading the same ten triples by hand
gives the opposite impression: the model did record a stance cue in $9/10$ notes, so
recording largely worked, and what was lost was lost later. The two $9/10$ figures are not
the same measurement, which is the problem: one counts sessions the gate flagged, the
other counts notes containing a cue. We report the pilot as unresolved and lean nothing
on it. The automatic scorer counts cue \emph{tokens}, and compression frequently
reworded a hedge rather than deleting it (``(unconfirmed)'' became ``pending
confirmation''), which that scorer records as a loss.
The open-class problem ate the measuring instrument inside a probe about the open-class
problem. $n=10$, one model.

\paragraph{Repairing the prose does not survive.} Gating the note and asking the model to
restore what it dropped produces a clean null end to end:
the repair fired on $8/10$ notes, cost $23.4\%$ more input tokens, and changed the final
verdict on zero of them, $0$ recovered and $0$ broken. The repair is not ignored; it is
undone. The model restores standing as a parenthetical (``(per informal report)'') and the
compressor removes it, which is what a compressor is for.

Reading those notes by hand produced the observation the rest of the paper is built on. A
few of the repaired notes happened to restore standing as a labelled line rather than as
a clause, and those survived compression $3/4$ times where the parentheticals survived
$0/4$. That is eight notes, and it is not evidence of anything: the notes that happened
to use fields differed from the notes that happened to use parentheticals in wording,
length and structure as well as in form. It is a hypothesis, and \S\ref{sec:twocell}
is the experiment that holds everything else constant to test it.

\paragraph{A third negative, in our own instrument.}
\label{sec:instrument}
The blind reader was asked for JSON. When a stored memory carried \emph{two} stance cues
it reliably emitted this:

\begin{promptbox}[what the judge actually returned]
\{"presented\_as": "qualified", "cue": "may have", "unconfirmed rumor from third party"\}
\end{promptbox}

which is not valid JSON. \texttt{json.loads} raised, the verdict was recorded as null,
and the scorer counted it as ``standing did not survive''. The instrument was discarding
its own clearest positives, and discarding them precisely on the memories with the most
surviving stance. It affected $6\%$ of the trials that existed when it was found. (That
is a different quantity from the $33$ of $4{,}104$ replies that need the fallback parser
today: one counts verdicts the bug destroyed before the fix, the other counts replies the
repaired parser recovers.) Because those trials were the
unambiguous ones, and because the parenthetical arm produced more of them, the bug
inflated the headline: on the affected data the pooled advantage read
$+9.3$ points, and re-running the identical four-level design after the fix gave $+4.0$.
(Both figures predate the two harder levels and the sixty-claim corpus, so neither is
comparable to the $+15.8$ and $+15.3$ of \S\ref{sec:twocell}; the comparison that matters
here is the one between the two of them.) We report this because
the direction is the instructive part. A measurement error in a paper's own scorer is
not a coin flip; it lands where the scorer is fragile, and a scorer is usually most
fragile on the cases its author considered obvious. The harness now stores every raw
judge reply and reports what fraction needed the fallback parser ($33$ of the
$4{,}104$ two-cell trials pooled across every corpus and model, with none
unusable), so the next failure of this kind is visible in the data rather than absorbed
by it.

\section{Field versus prose, held constant}
\label{sec:twocell}

\paragraph{Design.} Sixty claims, in seven registers, are each written twice, carrying
the \emph{identical} assertion and the \emph{identical} stance content, differing only in
where the standing sits:

\begin{promptbox}[the two cells]
\textbf{field}\quad Alice has admin access\\
\phantom{\textbf{field}}\quad SOURCE: a third party\\
\phantom{\textbf{field}}\quad CERTAINTY: rumor, unconfirmed\\[2pt]
\textbf{paren}\quad Alice has admin access (per a third party; rumor, unconfirmed)
\end{promptbox}

Both go to the same compressor with the same prompt and the same settings. Pairing is
strict: for a given (claim, pressure, replicate) the two arms receive the same filler
notes in the same positions, drawn from a seed that does not depend on the arm, so
nothing but form differs. Sampling is held fixed within every comparison. (Sonnet~5, which appears only in the
ten-claim runs and is not among the models reported here, rejects the temperature
parameter outright; see the Limitations.)

\paragraph{Pressure is a factor, not a nuisance.} A first version of this experiment
compressed a single one-sentence note and returned a clean null, for the good reason that
it did not compress: the input averaged $60$ characters and the stored memory came out at
$90$. Nothing was dropped because nothing had to be. Budget pressure is therefore swept:
\textsc{none} ($1$ note, ``one or two sentences''), \textsc{low} ($6$ notes, one line
each), \textsc{high} ($6$ notes, $40$ words total), \textsc{extreme} ($6$ notes, eight
words per line), \textsc{brutal} ($8$ notes, $25$ words total) and \textsc{severe} ($8$
notes, $12$ words total).

The last two levels were added after a first sweep in which the strongest model sat at
$100\%$ in both arms at every level, which we wrote up as ``no loss for the schema to
prevent''. That was the wrong conclusion: it had no loss to prevent \emph{at the
pressures we had applied}. A ceiling is a statement about the test, not about the
mechanism. The sixty-claim runs therefore use the two tight levels, where a compressor is
actually forced to discard something.

\paragraph{Readout.} A separate model sees only the stored memory and the claim to look
up, never the source note, the arm, or the pressure level, and answers whether the claim
is stored as established fact, as qualified, or is absent. Judge replies are parsed
without requiring valid JSON, for reasons given in \S\ref{sec:instrument}.

\begin{figure*}[t]
\centering
\includegraphics[width=\textwidth]{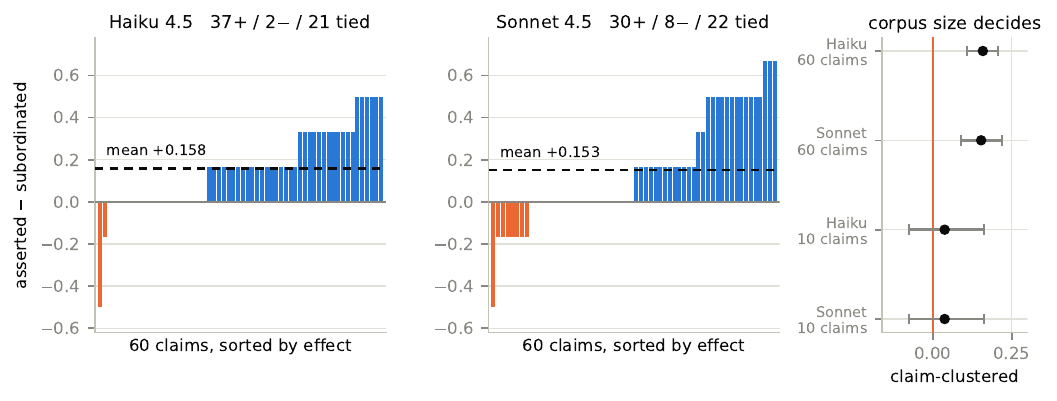}
\caption{\textbf{One bar per claim.} \textbf{Left and centre:} the field-minus-prose
difference in stance retention for each of the $60$ claims, sorted, at the two hard
budgets. Blue is a claim where the labelled field retained standing more often, orange
where the parenthetical did, grey where they tied. The dashed line is the mean.
\textbf{Right:} the same effect with claim-clustered $95\%$ intervals, for the $60$-claim
corpus and for the ten-claim corpus these models were previously measured on. The ten-claim
intervals contain zero; the sixty-claim intervals do not. Nothing about the models
changed between those two rows, only the claims they were asked about.}
\label{fig:perclaim}
\end{figure*}

\paragraph{Result.} Figure~\ref{fig:perclaim} and Table~\ref{tab:twocell}. On
\textbf{Haiku 4.5}, standing is recovered from $109/360$ field records against $52/360$
parentheticals, $30.3\%$ against $14.4\%$. On \textbf{Sonnet 4.5}, $202/360$ against
$147/360$, $56.1\%$ against $40.8\%$. The two models differ from each other by five and a
half thousandths.

\begin{table}[t]
\centering\small
\begin{tabular}{@{}l r r@{}}
\toprule
& \multicolumn{1}{c}{Haiku 4.5} & \multicolumn{1}{c}{Sonnet 4.5} \\
\midrule
field retains stance      & $30.3\%$  & $56.1\%$ \\
prose retains stance      & $14.4\%$  & $40.8\%$ \\
difference                & $+15.8$   & $+15.3$ \\
\addlinespace
claim-clustered $95\%$ CI & $[+10.8, +20.6]$ & $[+8.9, +21.9]$ \\
claims favouring field    & $37$      & $30$ \\
claims favouring prose    & $2$       & $8$ \\
claims tied               & $21$      & $22$ \\
sign test across claims   & $p<0.0001$ & $p=0.0005$ \\
\bottomrule
\end{tabular}
\caption{\textbf{Sixty claims, two models, matched design.} Rates are not comparable
across columns: Sonnet retains more standing overall because it is a better compressor at
these budgets. The comparison is within a column, and it is paired down to the individual
claim. The two models produce nearly the same difference. Differences are computed from
the raw counts and then rounded, so subtracting the two displayed percentages can differ
from the printed difference by a tenth. The Haiku difference is $109/360-52/360=57/360
=15.83\%$; subtracting the two rounded percentages instead gives $15.84$.}
\label{tab:twocell}
\end{table}

\paragraph{The claim is the unit that generalises.} A matched pair is not an independent
observation. Sixty claims run at two budgets with three replicates give $360$ pairs per
model, but a claim's particular wording recurs in six of them, so pair-level intervals
describe ``another replicate of these sixty claims'' rather than ``a claim we have not
written yet''. We therefore report the claim as the cluster throughout: a bootstrap that
resamples claims, and an exact sign test on the per-claim differences. Both are in
Table~\ref{tab:twocell}, and both are far more conservative than a pair-level test on the
same data, which we do not report because it treats correlated observations as
independent.

The sign test is the one we find most convincing, because it does not depend on effect
size at all. On Haiku the difference points toward the field form on $37$ claims and
toward prose on $2$. On Sonnet, $30$ against $8$. Whatever is happening is not carried by
a few lucky wordings.

\paragraph{Ten claims said the opposite, on the same models.} This experiment was first
run on ten claims, all in an office-work register. Restricted to the same two budgets and
the same two models, that corpus gives $+3.7$ points on Haiku, with per-claim differences
splitting four to five, and $+3.7$ on Sonnet, splitting three to four. Both
claim-clustered intervals contain zero. The earlier draft of this paper drew two
conclusions from that corpus, and both are wrong:

\begin{itemize}\itemsep2pt
\item that the effect \emph{disappears} once the claim is treated as the unit of
      analysis. It does not. It disappears on ten claims of one register, which is a
      statement about the corpus.
\item that the effect \emph{splits by model family}, being present on Haiku and Opus and
      exactly absent on both Sonnet models. It does not. Sonnet~4.5, measured across
      the full six-level ladder and then re-verified against an independent judge, moves
      from $+3.7$ points to $+15.3$ when it is asked about sixty claims instead of ten.
\end{itemize}

Both errors have the same cause and it is not subtle in hindsight. Ten stimuli in one
register is a sample of a genre, not a sample of claims, and it was small enough that
three wordings could carry a pooled average in either direction. It produced a false
null and a false moderator, and the false moderator was the more dangerous of the two
because it was interesting: a model-family split is the kind of finding that gets
written up rather than checked.

\paragraph{The judge is not the explanation.} In the runs above the compressor and the
blind reader are the same model, so each model in effect grades its own homework. That is
a live confound for any comparison \emph{between} models. We checked it by re-scoring, with
a single fixed judge, all $1{,}941$ memories that existed when that check was run, which
costs judge calls only and no new compression. That figure is a snapshot: it predates the
sixty-claim runs and the replication, so it is smaller than the $4{,}104$ trials the paper
now analyses. Agreement with the original scoring is $98\%$, and no model's effect moves by
more than two points; Sonnet~4.5's ten-claim null is unchanged to four decimals.

Two limits on what that buys, and they matter enough to state here rather than in the
Limitations. First, the fixed judge is Haiku~4.5, so it de-selfs the Sonnet column and
\emph{not} the Haiku one: Haiku memories were graded by Haiku in the original run and by
Haiku again in the check, and the headline sign split and the entire replication live in
that column. Second, both judges are Claude models, and agreement between two instruments
of one family measures variance within that family rather than bias shared across it. So
this is a check and not a removal. What carries weight against self-judging here is not the
rescore but the two instruments that are not model judges at all, the deterministic readout
and the hand labels in \S\ref{sec:errors}, both of which agree on direction. A non-Claude
rescore of the Haiku column is the obvious next control and we have not run it.

\paragraph{Which failure it prevents.} A claim can lose its standing or it can be
dropped from the digest entirely, and these are different failures. Under the
\textsc{brutal} budget on Haiku, the field form keeps standing $44\%$ of the time, is
stored as bare fact $11\%$, and is evicted $45\%$. The parenthetical keeps standing
$21\%$, is stored as bare fact $28\%$, and is evicted $52\%$. The eviction gap is small;
the washing gap is a factor of two and a half. On Sonnet, pooled over both budgets, the eviction rates are
statistically indistinguishable ($129$ against $123$ out of $360$, $p=0.52$) while the
stance difference is $15$ points.

That separation does not survive pooling on Haiku, and we report it because it narrows the
claim. Over both budgets the field form is evicted $199$ times out of $360$ against the
parenthetical's $228$ ($21$ against $50$ discordant pairs, exact McNemar $p=0.0008$), and
the gap is carried by \textsc{severe} ($p=0.006$) rather than by the \textsc{brutal} cell
the percentages above are drawn from ($p=0.07$). Decomposed, Haiku's $+15.8$-point stance
gain is $28$ fewer bare-fact stores plus $29$ fewer evictions, so about half of it runs
through retention rather than through qualification. The honest statement is therefore
model-conditional: on Sonnet, and on Haiku at \textsc{brutal}, the schema is not mainly
keeping claims in memory but keeping them qualified; on Haiku pooled it does both, in
roughly equal measure.

This also retracts a cost an earlier draft asserted, that the bulkier record is evicted
more often; see the Limitations.

\section{Where it holds and where it fails}
\label{sec:errors}

\paragraph{It is not one register.} The effect is positive on both models in six of the
seven registers. The strongest is \emph{project} ($+0.21$ on Haiku, $+0.23$ on Sonnet,
$17$ claims) and the weakest by a distance is \emph{conflict} ($+0.12$ on Haiku and effectively zero on
Sonnet,
$7$ claims). A corpus that was widened precisely because a narrow one had lied to us
should be checked for doing a smaller version of the same thing, and it is not.

\paragraph{An assumption-free test.} The sign test discards effect size and the cluster
bootstrap assumes its resampling distribution behaves. Shuffling the arm label within each
claim assumes neither. Over $20{,}000$ permutations the observed mean per-claim difference
is beyond every shuffled value on both models, $p=0.00005$, which is the resolution floor
of the test.

\paragraph{Ties are the tightest budget.} Pooling both budgets, $21$ of Haiku's $60$
claims tie and $22$ of Sonnet's do. Twelve of Haiku's $21$ are ties at zero, meaning
neither arm retained standing on any replicate. Splitting by budget shows where they come
from: at \textsc{severe} alone, $43$ of the $60$ claims tie, because both arms usually
lose the claim entirely. That dilutes the sign test rather than biasing it, and both
budgets remain significant separately (Haiku $p<0.0001$ and $p=0.0023$; Sonnet $p=0.011$
and $p=0.0005$).

\paragraph{The failure mode is disagreement.} The claims where the field form loses are
not scattered. Three of the six worst across the two models are claims whose sources
conflict:

\begin{promptbox}[the schema's blind spot]
CLAIM: The Helsinki office owns the relationship\\
SOURCE: sales said Helsinki, the account plan says Berlin\\
CERTAINTY: contested
\end{promptbox}

\noindent \texttt{SOURCE} and \texttt{CERTAINTY} are single slots. A claim with two
disagreeing attributions does not have one source, so the schema forces both back into
one field as a compressed clause, which is the same subordinate construction the schema
exists to avoid. The parenthetical carries the disagreement without strain because it was
never promising structure in the first place. A schema with room for more than one
attribution would be the obvious repair and we have not tested one.

\section{A pre-registered replication}
\label{sec:prereg}

Everything above shares a weakness that no re-analysis can remove. The fifty claims that
carry the result were written \emph{after} we had seen that the original ten produced a
null. That is an ordinary way for an experiment to develop and it is also exactly the
condition under which stimuli get tuned: having just learned that the corpus decides the
answer, we then wrote the corpus that produced the answer.

So we wrote sixty more claims to the same seven-register specification, together with a
prediction and a decision rule, and committed all of it to a public repository before any
call was made against them. The replication runs on Haiku~4.5 only, which is why
Table~\ref{tab:prereg} compares it against that model's numbers rather than the two-model
figure of the abstract. The registered rule was: a sign test at $p \ge 0.05$ means
the effect does not replicate and the headline claim becomes \textsc{not shown}; a
claim-clustered mean outside $[+8, +25]$ points means the size prediction failed, and we
say so whatever the sign test does. One run, three replicates, two budgets, no adding
levels or arms or replicates afterwards.

\begin{table}[t]
\centering\small
\begin{tabular}{@{}l r r@{}}
\toprule
& \multicolumn{1}{c}{original $60$} & \multicolumn{1}{c}{held-out $60$} \\
\midrule
field retains stance   & $30.3\%$ & $28.6\%$ \\
prose retains stance   & $14.4\%$ & $13.1\%$ \\
difference             & $+15.8$  & $+15.6$ \\
claim-clustered CI     & $[+10.8, +20.6]$ & $[+11.4, +20.0]$ \\
claims favouring field & $37$ & $38$ \\
claims favouring prose & $2$ & $1$ \\
\bottomrule
\end{tabular}
\caption{\textbf{The held-out replication.} Right-hand column: sixty claims written, and
a prediction and decision rule committed, before the run. The registered verdict is
\textsc{replicates}, and the size prediction held. The point estimate lands $0.2$ points
from the original. As in Table~\ref{tab:twocell}, the difference is computed before
rounding: $28.61-13.06=15.55$, shown as $+15.6$.}
\label{tab:prereg}
\end{table}

The result is in Table~\ref{tab:prereg}: $+15.6$ points, $38$ claims to $1$,
$p<10^{-6}$, none of the $720$ judge replies unparsable. By the registered rule this
replicates and the size prediction held.

We report the obvious caveat rather than let it pass. The new claims were written by the
same author, in English, to a specification we had already chosen; a genuinely
independent corpus would be stronger and was not available. What the exercise rules out
is narrower and still worth having: that the effect is an artefact of choosing stimuli
after seeing which ones worked.

\paragraph{A readout with no model in it.} Every outcome in this paper is one model's
judgement of another model's output, and the $98\%$ agreement between our two model
judges is agreement between two instruments of the same kind. We therefore also scored
the same stored memories with the deterministic gate of \citet{kwon2026factwash}, which
uses no model at all. It agrees on direction on both models and, on Haiku, on magnitude
to three decimals ($+0.158$ against the blind reader's $+0.158$; Sonnet $+0.311$ against
$+0.153$). The absolute rates are not comparable, because the gate only flags stance loss
it can detect and its recall on open-class stance plateaus near half. The agreement is
worth more than it first appears for a different reason: the gate is biased \emph{against}
the field form, since \texttt{CERTAINTY: hedged} contains no word its lexicon knows, and
the field form wins under it anyway.

\paragraph{And fifty items a human read.} Both readouts above are instruments. We
therefore hand-labelled a stratified sample of $50$ stored memories, drawn across both
models and both arms from $40$ distinct claims, using the same three-way question and the
same blinding the model judge got: the annotator sees the stored memory and the claim to
look up, and nothing about arm, model, pressure or claim id. The judge's verdicts were
held in a separate key file, and the scorer refuses a partially completed sheet.

Agreement is $86.0\%$, Cohen's $\kappa = 0.75$, on seven disagreements. Scoring those
same fifty items both ways gives $+20.8$ points by the judge and $+21.2$ by hand.

We drafted that margin as the judge understating the effect, which would have made the
paper's number conservative. Reading the disputed items withdrew it. All seven are printed
verbatim in Appendix~\ref{app:disputed}, and a third instrument was run over them:
content-word containment, the same no-model measure the companion gate uses to decide
whether a stored claim is matched to a source claim at all. On six of the seven the
claim's content is in the memory at or above the overlap the companion gate requires
before it will match a stored claim at all (containment $0.25$ to $0.83$, two of the six
sitting exactly on that threshold), so these are not the
judge matching text that is not there, which is the failure mode that would have
threatened the result. But in five of the seven the annotator answered \emph{absent} while
the content was present, and six of the seven sit in the prose arm, whose memories under
these budgets are dense comma-separated lists in which a four-word entry is easy to read
past. Annotator misses concentrated in one arm are exactly what produces a small
human-scored surplus in the other. The sign of those $0.4$ points is therefore
uninterpretable, and we withdraw the reading that it makes the judge conservative.

What survives is narrower and still worth having. The judge is not inventing claim
presence, and it agrees with a careful independent reading at $\kappa=0.75$. What does not
survive is any bound on the effect: the shift from changing scorer is $+0.3$ points with a
claim-clustered $95\%$ interval of $[-15.9, +16.0]$, which contains zero, so no shift was
detected, and which also fails to exclude a shift the size of the entire effect.
Disagreement is not spread evenly either, running $1/26$ on field records against $6/24$
on prose (Fisher exact $p=0.045$), and an asymmetry of that shape is the one that could
fabricate an effect even where, as here, it runs the other way.

And this is one rater, who is the paper's author: blind to condition, not blind to
hypothesis, with no second annotator and therefore no inter-rater agreement to report.
Printing all seven disputed items is the part of a second annotator we could supply
without one, since it lets a reader adjudicate them instead of taking our reading on
trust. It is a check on the instrument, not validation of it.

\section{Labels, layout, or length?}
\label{sec:ablation}

The comparison in \S\ref{sec:twocell} moves three things at once. The labelled record
names its parts, puts them on their own lines, and is slightly longer than the
parenthetical. Attributing the whole effect to the first of those, as an earlier draft of
this paper did, is not something the two-cell design can support.

The length check available from that data cannot settle it either, and it is worth saying
why, because the answer looked reassuring. Correlating each claim's effect against how
much longer its field form is returns $r=0.000$. That is not evidence of no confound: both
templates wrap the same three strings, so the difference is a constant twelve characters
for every claim and the predictor has no variance. A correlation of zero there is evidence
of no experiment.

\paragraph{Six forms, five contrasts.} We therefore ran four further arms on the
same sixty claims, the same two budgets, the same seed and the same replicate count. The
filler notes and the target's position are drawn from a generator that does not depend on
the arm, so every arm saw byte-identical surroundings and the already-collected field and
paren data could be reused rather than paid for again.

\begin{promptbox}[the six forms, mean length over the sixty claims]
\textbf{sentences} \hfill 120 chars\\
\quad X. This came from Y. It is Z.\\[2pt]
\textbf{field} \hfill 115 chars\\
\quad X / SOURCE: Y / CERTAINTY: Z\\[2pt]
\textbf{promoted} \hfill 103 chars\\
\quad X. Per Y; Z.\\[2pt]
\textbf{padded} \hfill 125 chars\\
\quad X (per Y; Z). Filed with the rest.\\[2pt]
\textbf{unlabelled} \hfill 101 chars\\
\quad X / from Y / Z\\[2pt]
\textbf{paren} \hfill 103 chars\\
\quad X (per Y; Z)
\end{promptbox}

\textsc{padded} is the control that does the work. It leaves the stance exactly where the
parenthetical puts it and appends a stance-free sentence, so it matches \textsc{sentences}
in token count and in having more than one sentence while carrying no additional
information about source, certainty or time. It is in fact the longest of the six, which
makes it a conservative test: if length were the mechanism, padding should have won
outright.

\begin{figure*}[t]
\centering
\includegraphics[width=\textwidth]{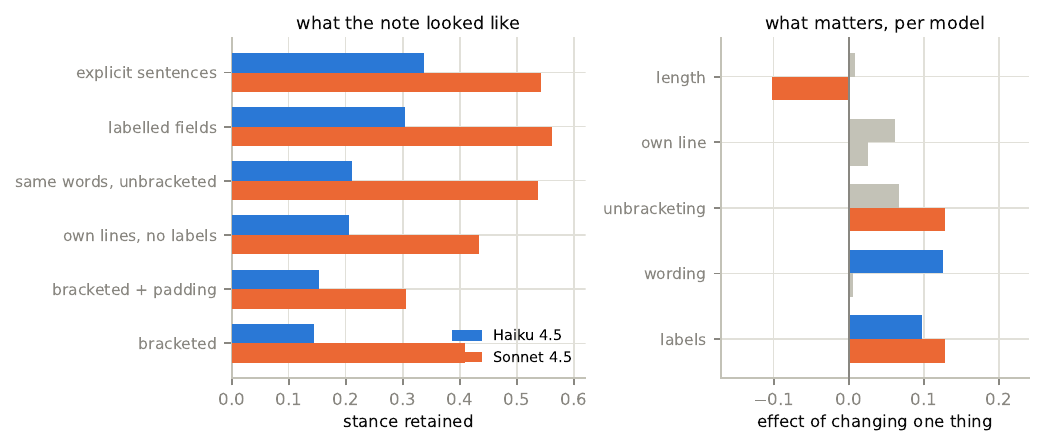}
\caption{\textbf{The same effect, assembled differently.} \textbf{Left:} share of stored
memories from which a blind reader can still recover the claim's standing, by the form the
stance was written in, for both models. Sonnet retains more overall because it is the
better compressor at these budgets; the comparison that matters is within a model.
\textbf{Right:} each row is one contrast between two of the forms on the left, named for
what differs between them: \emph{length} is padded against bracketed, \emph{own line} is
unlabelled against bracketed (which moves layout and bracketing together), \emph{
unbracketing} is promoted against bracketed, \emph{wording} is sentences against
promoted, and \emph{labels} is field against unlabelled. Grey bars are not significant at
$0.05$ by an exact sign test over per-claim differences. Only \emph{labels} is coloured
for both models; \emph{wording} carries the effect on Haiku and does nothing on Sonnet,
while \emph{unbracketing} does more on Sonnet. \emph{Length} is the one property that
helps neither model. Its apparent cost on Sonnet is the one bar here whose sign the
no-model readout reverses, so we do not claim it. Sixty claims, both hard budgets.}
\label{fig:ablation}
\end{figure*}

\paragraph{Result.} Figure~\ref{fig:ablation} and Table~\ref{tab:ablation}. We ran the
full set of arms on both models, and the comparison is the finding: the two models reach
the \emph{same} net effect by \emph{different} routes.

\begin{table}[t]
\centering\small
\begin{tabular}{@{}l r r@{}}
\toprule
what changes & Haiku 4.5 & Sonnet 4.5 \\
\midrule
labels                & $+9.7^{*}$  & $+12.8^{*}$ \\
unbracketing          & $+6.7$      & $+12.8^{*}$ \\
wording it explicitly & $+12.5^{*}$ & $+0.6$ \\
own line              & $+6.1$      & $+2.5$ \\
length                & $+0.8$      & $-10.3^{*}$ \\
\midrule
net (field vs bracketed) & $+15.8^{*}$ & $+15.3^{*}$ \\
\bottomrule
\end{tabular}
\caption{\textbf{The same effect, assembled differently, under the blind reader.}
Percentage-point change in
stance retention from altering one property, per claim, sixty claims, both hard budgets.
$^{*}$ marks a significant exact sign test across claims. The net effect at the bottom is
near-identical on the two models; almost none of the components are. The \emph{length}
row is the one whose sign the no-model readout reverses ($-1.9$ on Haiku, $+5.8$ on
Sonnet), so its $-10.3$ is a property of this instrument and we claim only the null.}
\label{tab:ablation}
\end{table}

Three things survive the second model and one does not.

\begin{itemize}\itemsep2pt
\item \textbf{Labels replicate.} $+9.7$ points $[+2.8, +16.1]$ on Haiku and $+12.8$
      $[+7.2, +18.3]$ on Sonnet, both significant, intervals overlapping. This is the
      only component that behaves the same way on both models. We have been wrong about
      labels in both directions: an early draft credited them with the whole effect, a
      later one dismissed them as irrelevant, and the measurement puts them second of
      five and reproducible.
\item \textbf{Length helps on neither model.} $+0.8$ $[-2.5, +3.9]$ on Haiku is a
      tight null; on Sonnet the same padding scores $-10.3$ points
      $[-15.0, -5.3]$, $p=0.0001$ under the blind reader. We claim the null and not the
      harm: the no-model readout below puts this contrast at $-1.9$ on Haiku and
      $+5.8$ on Sonnet, so it is the one component whose \emph{sign} depends on which
      instrument is asked, and \emph{padding is actively harmful} is not a finding two
      instruments support. Making the note longer without making the stance plainer buys
      nothing; whether it costs is not settled here.
\item \textbf{Unbracketing helps on both, by different amounts}, $+6.7$ and $+12.8$; it
      reaches significance on Sonnet and not on Haiku. Note that on Haiku \emph{own line}
      ($+6.1$) is slightly smaller than \emph{unbracketing} ($+6.7$) even though it moves
      layout and bracketing together, which implies the layout part alone is marginally
      negative there. The difference is well inside both intervals and we do not read
      anything into it.
\item \textbf{Wording does not replicate.} It is the largest component on Haiku
      ($+12.5$, $p=0.0001$) and worth nothing at all on Sonnet ($+0.6$, $p=0.77$, with a
      $95\%$ interval of $[-6.1, +7.2]$ that comfortably contains zero).
\end{itemize}

\noindent So a single-model ablation would have licensed the wrong summary in either
direction. Run only on Haiku, this section would say the effect is mostly about how
explicitly stance is worded. Run only on Sonnet, it would say wording is irrelevant and
the whole thing is brackets. Both are what one model shows and neither is what two models
show together. The reportable statement is narrower than either: \emph{labels help on both
models, length helps on neither, and which of the remaining properties carries the effect
depends on the model.}

\paragraph{The same contrasts, scored by an instrument with no model in it.} Everything
above is one Claude-family judge reading Claude-family output, which is the objection
\S\ref{sec:errors} answers for the two-cell result and did not, until now, answer here.
\texttt{experiments/ablation\_mechanical.py} reruns all seven contrasts with the
deterministic gate substituted for the blind reader. Read the magnitudes as floors: the
gate has a documented recall ceiling on open-class stance and is tilted against any arm
carrying stance as a label, so it cannot see \texttt{CERTAINTY: rumor} at all. Five of the
seven contrasts keep their direction on both models, and the two largest, subordination
and the whole asserted path, come out \emph{larger} under the no-model instrument
($+11.7$ and $+16.9$ on Haiku, $+26.9$ and $+30.6$ on Sonnet). Two do not survive. Length
reverses sign on both models, which is why we claim the null and not the harm above. The
labels contrast reads $-1.1$ on Haiku and $+0.3$ on Sonnet, which is the predicted
consequence of an instrument blind to the label form rather than evidence against labels,
and we say so rather than resolve it: on our own ledger's vocabulary, a component that
only one instrument can see is not shown by two. The full table ships in
\texttt{data/results/ablation\_mechanical.json}.

\paragraph{What this means.} A compressor asked to discard material does not decide by
volume. That much holds on both models and is the firmest thing here: padding buys
nothing on either, and the one reading on which it actively costs does not survive a
second instrument. What it does respond to is some property
of how plainly the stance is presented, and the two models weight the available ways of
being plain differently. Naming the parts with labels works on both. Getting the words out
of the brackets works on both, more on Sonnet. Saying the same thing in fuller sentences
works on Haiku and does nothing on Sonnet.

We are deliberately not naming a single mechanism. The evidence for one would have to look
like agreement across models about which property matters, and what we have is agreement
about the outcome with disagreement about the route.

It also accounts for a failure mode we had already found, before this ablation existed.
The claims where the field form loses are the ones whose sources disagree
(\S\ref{sec:errors}): \texttt{SOURCE} and \texttt{CERTAINTY} are single slots, so ``sales
said Helsinki, the account plan says Berlin'' has to be crushed back into one field, which
returns the stance to a subordinate blob. Two analyses that did not know about each other
point at the same mechanism.

\paragraph{What survives for the schema.} The comparator here is the \textsc{sentences}
arm, which states the source and the certainty in full sentences; we call it \emph{explicit
prose} rather than ``asserted prose'', because elsewhere in this paper a claim stored as
\emph{asserted} is one whose standing has been stripped, which is the opposite of what this
arm does. We could not detect a difference between the schema and explicit prose on either
model, and the interval is too wide to call them equivalent. What we can say is that the
schema is not the only way to buy retention. What the schema still has, and what this
ablation does not touch, is that its retention is \emph{checkable}. A four-field record can
be compared against its source by a program with no model and no lexicon of hedged
phrasings (\S\ref{sec:check}). Prose that retains stance has to be read by something that
can recognise stance in arbitrary wording, which is the open-class problem that plateaus
near half recall \citep{kwon2026factwash}. Retention and verifiability were conflated in
earlier drafts of this work; they are separate, and only one of them is about labels.

\section{The schema end to end}
\label{sec:endtoend}
The two-cell experiment isolates form. It does not tell us whether a model asked to
\emph{use} the schema does so well enough for the isolated effect to appear in a
pipeline. A pilot suggests it can, on the same model where the two-cell effect was
found: writing the note to the schema and then compressing to ordinary prose raised
clean writes from $3/10$ to $7/10$, and when the model filled the schema it filled it
completely, keeping all four fields in $10/10$ records. It classified $8/10$ sources as
something other than plainly asserted ($6$ rumor, $2$ hedged), so the recording step
works. This is $n=10$ on one model and is a pilot, not a result.

\paragraph{Independent readout.} The pilot above is scored by a deterministic checker we
also wrote, which is not an independent judge of our own intervention. The two-cell
experiment therefore uses a different readout entirely: a model that sees only the stored
memory and the claim to look up, never the source, the arm, or the pressure level. That
is the readout every number in \S\ref{sec:twocell} comes from, and it is the reason we
trust the null on Sonnet as much as the positive on Haiku.

\section{Checking a record}
\label{sec:check}

Once standing is a field, checking it stops needing a model. The hard half of detection
is recognising stance in a rewrite, an open-class problem with no finite vocabulary. A
field does not need recognising. The released check compares the record against the
source with three rules: if the source hedges, \texttt{CERTAINTY} may not read
\texttt{asserted}; if the source names a speaker, \texttt{SOURCE} may not be empty; if
the source dates the claim, \texttt{AS\_OF} may not be empty. A record missing its
retention fields is \textsc{malformed}, never a pass.

\paragraph{An instrument note.} A prose detector scores a schema memory
\emph{worse} than a prose memory, because \texttt{CERTAINTY: hedged} preserves the
standing perfectly and contains no word a hedge lexicon knows.
Measured: a schema note compressed into a schema memory scored $5/10$ clean under the
prose detector against $7/10$ for the same schema note compressed into prose, an
apparent regression that is entirely an artefact of the instrument. Inspection of the
records showed all four fields intact in $10/10$ cases. The record is more
machine-readable, not less; it has to be read as a record.

\section{Conclusion}

Write a claim's standing into a bracketed aside and a compressor treats it as an aside.
Write the same standing as a labelled field and it survives about fifteen points more
often, across sixty claims in seven registers, on two models whose estimates differ by five and a half thousandths, and again on sixty further claims written before the run under a
prediction registered in advance.

What we cannot do is tell you why in one sentence, and the reason is worth more than the
sentence would have been. Ablating the format on both models gives the same net effect
assembled from different parts. Labels help on both. Length helps on neither, and whether
padding a note with stance-free text actively costs depends on which instrument is asked,
so we report the null and not the harm. But the largest single
component on Haiku, wording the stance as a full sentence, is worth nothing at all on
Sonnet. Either model on its own would have supported a confident mechanism, and the two
mechanisms would have contradicted each other.

So the advice is the intersection rather than the theory: state the standing explicitly,
do not settle for making the note longer, and if you are choosing a write format for a
particular model, measure it on that model. The labelled schema keeps one advantage this
experiment cannot take from it, which is that a program can check whether its fields are
there at all; prose that retains stance still has to be read by something that can
recognise stance in arbitrary wording.

Most of what we learned here arrived as negatives, and the positive result exists because
they were caught. A ten-claim corpus in one register manufactured both a false null and a
false model-family split, and we believed the second long enough to verify it against an
independent judge and find the check reassuring. Our own scorer discarded its clearest
verdicts and inflated an effect twofold. A single-model ablation named a mechanism that
the second model declined to confirm. Every number above is stated at the level the last
surviving control allows, and the ledger in Appendix~\ref{app:claims} lists the nine
claims we have withdrawn.

\section*{Limitations}

\paragraph{Scope of the models.} Two models from one provider is not a sample of
compressors. Both were run through the full ablation, which is what let us discover that
the decomposition does not transfer; with a third model it might not even be two groups.
We are wary of claims about which models do or do not behave a given way, because an
earlier and smaller version of this work produced a confident model-family split that
turned out to be an artefact of its stimuli. Opus~4.5 and Sonnet~5 were measured only on
the ten-claim corpus and are therefore not reported as results.

\paragraph{Scope of the stimuli.} Sixty hand-written English claims across seven
registers, paired with stance-free fillers and constructed so the two arms could be
matched exactly. That control is what makes the comparison clean and also what makes it
artificial: real notes are not matched pairs, and a real memory system writes its own
claims rather than receiving ours. Roughly a third of claims tie on each model, mostly
because at the tightest budget both arms lose the claim entirely; ties carry no
information for the sign test, so the reported $p$-values are conservative in that
respect and say nothing about those claims. The padding sentence in the ablation is a
single fixed string, though it is hard to see how a different stance-free sentence could
achieve more than the longest arm already failed to achieve.

\paragraph{Scope of the conditions.} The two budgets reported here are the tight end of a
six-level ladder. At looser budgets the effect is small and we do not claim it. The honest
statement is that this matters when a memory system is short of room, which is the usual
condition but not the only one. ``Standing recovered'' is also one model's judgement of
another model's output: blind to the condition, reproduced by a single fixed judge, and
checked against $50$ hand labels ($\kappa=0.75$, \S\ref{sec:prereg}) that are themselves
the author's and whose disagreements are printed rather than summarised.

\paragraph{The ablation is a chain, not a factorial, and it does not transfer.} The six
forms move one thing at a time along a path; they do not independently cross labelling,
bracketing, wording and length, so interactions are unmeasured and the component sizes are
a ranking under this particular chain rather than coefficients that would survive a full
crossing. More importantly, the ranking is model-specific: run on Haiku alone it says the
effect is about wording, run on Sonnet alone it says wording is irrelevant. Two models is
enough to show the decomposition does not transfer and not enough to say what governs it.

\paragraph{One human read fifty of these, and that human wrote the paper.} Outcomes are
model judgements, checked against a fixed second judge ($98\%$), a deterministic gate, and
$50$ hand labels ($\kappa=0.75$, \S\ref{sec:prereg}) whose seven disagreements are printed
in Appendix~\ref{app:disputed} rather than summarised away. The annotator was blind to condition but is the author, so blind
to condition is not blind to hypothesis, and one rater yields no inter-annotator
agreement. The sample also cannot bound the thing it was drawn to check: the interval on
the scorer-swap shift spans about the size of the effect. Reading the disputed items found
the annotator, not the judge, to be the one missing claims, which is a reason to trust the
judge slightly more and the hand labels rather less. The missing control is a larger
blinded annotation by people with no stake in the result, reported with inter-rater
agreement. We could not obtain one, and fifty of the author's own labels do not stand in
for it.

\paragraph{What the schema still cannot do.} Every failure measured here is a field being
\emph{dropped}. A field confidently filled with the wrong value, \texttt{CERTAINTY:
asserted} on a rumour, is a worse failure and is invisible to a checker that only asks
whether the field is present; ours catches the subset where the source's own hedging is
detectable. Detecting stance in the \emph{source} still uses deterministic English
lexicons, with every open-class limit that implies. And a single \texttt{SOURCE} slot
cannot hold two disagreeing attributions, which is the failure mode of
\S\ref{sec:errors}; a schema with room for more than one attribution is the obvious
repair and we have not tested it.

\paragraph{A concern we raised and then undercut ourselves.} An earlier draft argued that
a multi-line record costs budget a compressor has to find somewhere, and that this should
show up as field records being evicted more often. It did, once, on one model at one
ladder length. It survives neither the full ladder nor the larger corpus, and the ablation
then removed its premise: length does not drive retention at all, so ``the record is
bulkier'' is not obviously a cost. We leave the concern recorded rather than deleted,
because a schema with many more fields than four would test it again.

\paragraph{Cost.} The end-to-end pilot ($n=10$, one model) is a pilot and nothing in the
conclusion rests on it. Metered API spend for everything reported here is under fourteen
dollars.

\section*{Ethics Statement}

All stimuli are synthetic, written by the author for this experiment. They describe
fictional colleagues and fictional internal facts, contain no personal data, and were not
drawn from any real correspondence. The failure this work addresses is one where a system
overstates its confidence to its own operator; the intervention is a change to how a
system records what it was told, and we see no dual-use concern in releasing it. Raw runs
including every model output are released alongside the code so the analysis can be
checked rather than taken on trust.

\bibliography{references}

\appendix
\clearpage
\onecolumn

\noindent\begin{minipage}{\textwidth}
\subsection*{Appendix contents}
\noindent
\hyperref[app:claims]{\textbf{A}\quad Claims and evidence} \dotfill \pageref{app:claims}\\
\hyperref[app:stimuli]{\textbf{B}\quad Stimuli and the pressure ladder} \dotfill \pageref{app:stimuli}\\
\hyperref[app:disputed]{\textbf{C}\quad Every item the human and the judge disagreed on} \dotfill \pageref{app:disputed}\\
\hyperref[app:repro]{\textbf{D}\quad Reproducibility} \dotfill \pageref{app:repro}
\end{minipage}
\bigskip

\section{Claims and evidence}
\label{app:claims}

Every load-bearing claim, its evidence, and its epistemic status. Same discipline and
same format as the companion paper \citep{kwon2026factwash}.

\begin{itemize}\itemsep2pt
\item \textsc{shown}: direct measurement supports it.
\item \textsc{retracted}: \emph{we} asserted it earlier and later withdrew it. Three of
      the nine were at some point this paper's \textbf{title claim}, and are marked
      \textsc{(headline)}.
\item \textsc{not shown}: our measurement neither supports nor refutes it.
\item \textsc{narrowed}: we asserted it and it survived, but smaller or under fewer
      conditions than first stated.
\item \textsc{not claimed}: we never asserted it; the row exists so a reader cannot
      infer it.
\item \textsc{pending}: this draft still owes the measurement.
\end{itemize}

\medskip
{\small
\begin{longtable}{@{}L{0.46\textwidth} L{0.18\textwidth} L{0.28\textwidth}@{}}
\caption{\textbf{Claims and evidence.}}\label{tab:claims}\\
\toprule
\textbf{Claim} & \textbf{Evidence} & \textbf{Status} \\
\midrule
\endfirsthead
\multicolumn{3}{@{}l}{\emph{Table~\ref{tab:claims}, continued}}\\
\toprule
\textbf{Claim} & \textbf{Evidence} & \textbf{Status} \\
\midrule
\endhead
\midrule \multicolumn{3}{r@{}}{\emph{continued on next page}}\\ \endfoot
\bottomrule \endlastfoot

Making the note longer retains more stance. & \S\ref{sec:ablation},
Tab.~\ref{tab:ablation} & \textsc{not shown, on two models}: $+0.8$ points
$[-2.5,+3.9]$ on Haiku and $-10.3$ $[-15.0,-5.3]$ on Sonnet under the blind reader \\

\textbf{Padding a note actively costs stance on Sonnet.} & \S\ref{sec:ablation} &
\textsc{narrowed}: asserted from the blind reader's $-10.3$. The no-model readout puts
the same contrast at $+5.8$ on Sonnet and $-1.9$ on Haiku, so it is the one component
whose sign depends on the instrument, and the null is all we claim \\

Labels contribute to retention. & \S\ref{sec:ablation} & \textsc{shown on both models}:
$+9.7$ $[+2.8,+16.1]$ and $+12.8$ $[+7.2,+18.3]$. The only component that behaves the
same way on both \\

\textbf{The ablation identifies the mechanism.} & \S\ref{sec:ablation} &
\textsc{not claimed}: the two models agree on the net effect and disagree about which
property produces it. Wording is the largest component on Haiku and worth nothing on
Sonnet \\

\textbf{Labels are not the mechanism.} \textsc{(headline)} & \S\ref{sec:ablation} & \textsc{retracted twice,
in opposite directions}: first asserted, then over-corrected to dismissal, and finally
measured on a second model where labels are the \emph{most} robust component \\

\textbf{Our consistency gate protects this paper from stating what it has withdrawn.} &
\S\ref{app:repro} & \textsc{retracted}: it re-derived every number from the run data and
caught none of six internal contradictions found by a reading, because each was a
sentence surviving a rewrite rather than a number drifting. The introduction asserted a
mechanism the paper retracts in \S\ref{sec:ablation}. The gate now also checks prose
against prose \\

The field-over-bracketed effect replicates on claims chosen before the result was known.
& \S\ref{sec:prereg}, Tab.~\ref{tab:prereg} & \textsc{shown}: pre-registered at a public
commit before any call, $+15.6$ points $[+11.4,+20.0]$, $38$ claims to $1$, size
prediction held \\

The effect survives an instrument with no model in it. & \S\ref{sec:prereg} &
\textsc{shown}: a deterministic gate agrees on direction for both models and on magnitude
to three decimals for Haiku, while being biased \emph{against} the field form \\

\textbf{Grammatical subordination is the mechanism.} \textsc{(headline)} &
\S\ref{sec:ablation} & \textsc{retracted}: asserted by a previous draft at $+18.3$ points,
a figure from the four-level ladder before the sixty-claim corpus. On the current data the
same path is $+19.2$, and it splits into $+6.7$ for unbracketing and $+12.5$ for wording;
neither figure sums to the $+15.8$ net, which walks a different path \\

Which property retains stance is the same across models. & \S\ref{sec:ablation},
Fig.~\ref{fig:ablation} & \textsc{not shown}: the two models agree on the net
effect and disagree on its composition. Wording is the largest component on
Haiku ($+12.5$) and null on Sonnet ($+0.6$); unbracketing runs the other way \\

The judge manufactures part of the effect. & \S\ref{sec:prereg},
\S\ref{app:disputed} & \textsc{not shown}: on the seven items where a human reader
disagreed, six memories contain the claim's content by a no-model containment check, so
the judge is not matching text that is not there. Agreement $86\%$, $\kappa=0.75$ \\

The judge understates the effect, so the paper's number is conservative. &
\S\ref{sec:prereg} & \textsc{retracted}: drafted from $+21.2$ by hand against $+20.8$ by
the judge. Five of the seven disagreements are the annotator answering \emph{absent} where
the content was present, and six of seven are in one arm, which is what produces a surplus
of that size in the other \\

\textbf{Fifty hand labels validate the model judge.} & \S\ref{sec:prereg} &
\textsc{not claimed}: one rater, who is the paper's author, blind to condition but not to
hypothesis, with no inter-rater agreement. The scorer-swap interval $[-15.9,+16.0]$ does
not exclude a shift the size of the effect \\

The judge errs evenly across the two arms. & \S\ref{sec:prereg} & \textsc{not shown}:
disagreement runs $1/26$ on field and $6/24$ on prose, Fisher exact $p=0.045$, and the
errors that touch the difference cancel by arithmetic rather than by design \\

The effect is a matter of length or token count. & \S\ref{sec:ablation} &
\textsc{not shown}: the longest arm in the experiment, a parenthetical padded with
stance-free text, is worth $+0.8$ points, $95\%$ CI $[-2.5,+3.9]$. The interval is narrow
as well as containing zero, so this one can be called negligible \\

\textbf{The labelled schema performs as well as explicit prose (the \textsc{sentences}
arm).} &
\S\ref{sec:ablation} & \textsc{not claimed}: no difference detected on either model
($-3.3$ points on Haiku, $p=1.0$; $+1.9$ on Sonnet), but Haiku's $90\%$ interval
$[-9.2,+1.9]$ fails a two-one-sided test against a $5$-point margin. Absence of a detected
difference is not equivalence \\

Stance is retained in proportion to how plainly it is stated. & \S\ref{sec:ablation} &
\textsc{shown in direction, not in composition}: every property that makes the stance
plainer helps on at least one model, and no property helps equally on both \\

A labelled record's retention can be verified without a model. & \S\ref{sec:check} &
\textsc{shown} by construction, and it is the schema's remaining advantage over prose,
which this ablation does not touch \\

The schema handles claims whose sources disagree. & \S\ref{sec:errors} &
\textsc{not shown}: those are where it loses. One \texttt{SOURCE} slot cannot hold two
attributions, so they are compressed back into a subordinate clause \\

Stance written as a labelled field survives compression more often than the same stance
written as prose. & \S\ref{sec:twocell}, Fig.~\ref{fig:perclaim} & \textsc{shown} on
$60$ claims, two models: Haiku $+15.8$ points, Sonnet $+15.3$; claim-clustered
$p<0.001$ on both; $37$ vs $2$ and $30$ vs $8$ claims by sign \\

The effect is a property of claims in general, not of a few wordings. &
\S\ref{sec:twocell} & \textsc{shown}: the sign test over per-claim differences is the
headline test, and it does not depend on effect size \\

The mechanism is that standing is washed off, not that the claim is dropped. &
\S\ref{sec:twocell} & \textsc{shown for Sonnet and for Haiku at the brutal budget}: prose
is stored as bare fact $28\%$ vs the field's $11\%$ while eviction is $52\%$ vs $45\%$,
and on Sonnet eviction is indistinguishable ($p=0.52$). \textsc{Narrowed} on Haiku pooled
over both budgets, where the field form is also evicted less ($199$ vs $228$ of $360$,
McNemar $p=0.0008$) and about half the stance gain runs through retention \\

The effect requires the compressor to be under budget pressure. & \S\ref{sec:twocell} &
\textsc{shown for the tight end of the ladder}; at loose budgets it is small and we do
not claim it \\

\textbf{The effect disappears once the claim is the unit of analysis.} &
\S\ref{sec:twocell} & \textsc{retracted}: asserted on a ten-claim corpus. The clustering
correction was right; the conclusion drawn from it was an artefact of ten stimuli in one
register \\

\textbf{The effect splits by model family, being absent on Sonnet.} &
\S\ref{sec:twocell} & \textsc{retracted}: asserted on the full ten-claim ladder and
re-checked against an independent judge, which agreed. Sonnet~4.5 goes from $+3.7$ to
$+15.3$ points on sixty claims \\

\textbf{The schema helps most where the compressor is weakest.} \textsc{(headline)} & \S\ref{sec:twocell} &
\textsc{retracted}: claimed by an earlier draft on a shorter pressure ladder \\

\textbf{The schema costs retention: a bulkier record gets evicted more.} &
\S\ref{sec:twocell} & \textsc{retracted}: shown once on one model at one ladder length
($8$ vs $1$, $p=0.039$); survives neither the full ladder nor the larger corpus \\

The advantage widens monotonically as pressure increases. & \S\ref{sec:twocell} &
\textsc{retracted}: predicted in the harness before the first run, not observed \\

Opus~4.5 and Sonnet~5 show the effect. & \S\ref{sec:twocell} & \textsc{not claimed}:
both were measured only on the ten-claim corpus, which we no longer trust for this
question \\

A fixed independent judge changes the picture. & \S\ref{sec:twocell} &
\textsc{not shown}: $1{,}941$ memories re-scored by one judge, $98\%$ agreement, no
model's effect moves more than two points \\

A write schema raises the share of memory writes that retain standing. &
\S\ref{sec:endtoend} & \textsc{pending}: pilot $3/10 \to 7/10$, one model, one prompt
set, $n=10$ \\

The loss occurs at recording rather than at compression. & \S\ref{sec:negatives} &
\textsc{pending}: pilot $9/10$ vs $1/10$ on one model \\

Our blind reader measured what we thought it measured. & \S\ref{sec:instrument} &
\textsc{not shown for the first runs}: a parser bug discarded $6\%$ of verdicts,
non-randomly, and inflated the effect from $+4.0$ to $+9.3$ points. Fixed, re-run, and
the pre-fix data is archived rather than pooled \\

Repairing the note in prose fixes the stored memory. & \S\ref{sec:negatives} &
\textsc{not shown}: clean null in the pilot, $0$ recovered and $0$ broken \\

Field-level checking needs no model on the memory side. & \S\ref{sec:check} &
\textsc{shown} by construction; the released checker uses no model \\

\textbf{This paper's detector is an independent judge of this paper's intervention.} &
\S\ref{sec:endtoend} & \textsc{not claimed}: it is our own gate, which is why a second
readout is required \\

\end{longtable}
}

\section{Stimuli and the pressure ladder}
\label{app:stimuli}

Sixty claims. Each is written twice; both forms carry the same assertion and the same
stance content, and differ only in where the standing sits. The claim text itself is a
bare assertion containing no stance, so all the stance lives in the source and certainty
fields and the two arms can be matched exactly.

\medskip
{\small
\begin{tabular}{@{}l r L{0.56\textwidth}@{}}
\toprule
\textbf{Register} & \textbf{$n$} & \textbf{Why it is here} \\
\midrule
project     & 17 & the original register; long noun phrases, named colleagues \\
medicine    &  8 & clinical detail and numbers a compressor is reluctant to drop \\
news        &  7 & named external actors, which invite attribution by default \\
personal    &  7 & first-person recall, where the source \emph{is} the speaker's memory \\
scheduling  &  7 & dates and times, the one field type with its own retention slot \\
estimate    &  7 & numeric quantities carrying their own implicit precision \\
conflict    &  7 & two sources disagreeing, so ``who said it'' is load-bearing \\
\midrule
total       & 60 & \\
\bottomrule
\end{tabular}
}

\medskip
The first ten (all \emph{project}) are the corpus this experiment originally ran on and
are kept as a distinct subset rather than merged, so the earlier runs stay comparable and
the corpus effect of \S\ref{sec:twocell} can be measured rather than assumed. Two
examples, one from each end of the register range:

\begin{promptbox}[two of the sixty]
\textbf{field}\quad The patient is allergic to penicillin\\
\phantom{\textbf{field}}\quad SOURCE: the patient's adult son\\
\phantom{\textbf{field}}\quad CERTAINTY: reported, not confirmed against records\\[3pt]
\textbf{paren}\quad There were 18 affected accounts (per support said 18,\\
\phantom{\textbf{paren}}\quad the data team said 25; conflicting figures)
\end{promptbox}

Filler notes are deliberately stance-free and mundane (``Standup moved to 9:45am on
Tuesdays'', ``Priya owns the billing service''). They exist to consume budget. For a
given (claim, pressure, replicate) both arms receive the same fillers in the same
positions, drawn from a seed that does not depend on the arm, so the fillers cannot be
what differs. The full corpus is released as a single Python file.

\medskip
\noindent\textbf{The pressure ladder.} Each level names how many notes go in and what
budget comes out.

\medskip
{\small
\begin{tabular}{@{}l r L{0.55\textwidth}@{}}
\toprule
\textbf{Level} & \textbf{Notes} & \textbf{Budget given to the compressor} \\
\midrule
\textsc{none}    & 1 & one or two sentences \\
\textsc{low}     & 6 & one short line per fact \\
\textsc{high}    & 6 & at most 40 words total \\
\textsc{extreme} & 6 & one line per fact, maximum 8 words per line \\
\textsc{brutal}  & 8 & at most 25 words total \\
\textsc{severe}  & 8 & at most 12 words total, keep only what matters most \\
\bottomrule
\end{tabular}
}

\medskip
The sixty-claim runs use \textsc{brutal} and \textsc{severe}, the tight end, because that
is where a compressor is actually forced to discard something. The four looser levels
were run on the ten-claim corpus and are reported in \S\ref{sec:twocell} only as the
condition under which the effect is small.

\section{Every item the human and the judge disagreed on}
\label{app:disputed}

Fifty stored memories were labelled by hand, blind to arm and model, and the judge
disagreed on seven. Rather than report an agreement statistic and ask to be believed about
whether those seven are benign, all of them are below: the claim the annotator was asked
to look up, the memory exactly as it was stored, both labels, and the content-word overlap
between claim and memory. That last column is a no-model string measure, the same
containment the companion gate uses to decide whether a stored claim is matched to a
source claim at all, so it is a third instrument on this subset rather than one invented
to settle the question.

Six of the seven memories carry the claim's content, so the disagreements are about what
standing that content is stored with, not about whether the judge matched text that is not
there. Five of the seven are the annotator answering \emph{absent} where the content is
present. Seven items cannot support a statistic and none is offered; the point of printing
them is that a reader can disagree with both labellers.

\medskip
\input{disputed_table}

\section{Reproducibility}
\label{app:repro}

\paragraph{Models.} The results reported here use Claude Haiku~4.5 and Sonnet~4.5.
Opus~4.5 and Sonnet~5 were also run, on the ten-claim corpus only, and are not reported
as results. Within a run the compressor and the blind reader are the same model; because
that makes each model its own grader, we additionally re-scored every stored memory with
a single fixed judge (Haiku~4.5), which reproduces the same picture at $98\%$ agreement.
The compressor samples at temperature $1.0$ and the reader at $0.0$, except on Sonnet~5,
which rejects the temperature parameter outright and therefore sampled at its default;
its summaries record this.

\paragraph{Where the $4{,}104$ trials come from.} The denominator quoted for judge-parse
health is every scored trial in the project, not one experiment, and it is worth breaking
out because it cannot be reconstructed from the headline design alone:

\medskip
{\small
\begin{tabular}{@{}l l r@{}}
\toprule
corpus & model & trials \\
\midrule
ten-claim ladder & Haiku 4.5   & 640 \\
                 & Opus 4.5    & 536 \\
                 & Sonnet 4.5  & 480 \\
                 & Sonnet 5    & 288 \\
\addlinespace
sixty-claim      & Haiku 4.5   & 720 \\
                 & Sonnet 4.5  & 720 \\
\addlinespace
held-out replication & Haiku 4.5 & 720 \\
\midrule
\multicolumn{2}{@{}l}{total} & 4{,}104 \\
\bottomrule
\end{tabular}
}

\medskip
The ten-claim rows are uneven because those runs covered different subsets of the pressure
ladder as the budget allowed, and because incomplete pairs are dropped rather than
averaged. The $1{,}440$ ablation trials per model are scored separately and are not in
this table.

\paragraph{What is released.} The repository is
\url{https://github.com/collapseindex/factwash} (Apache-2.0). It contains the harness
(\texttt{experiments/two\_cell\_pressure.py}), the scorer
(\texttt{experiments/two\_cell\_analyse.py}), the figure script, this paper's source, and
\emph{every raw run}: one JSON line per trial carrying the note, the stored memory, the
blind reader's raw reply, and how that reply was parsed. Superseded runs are published
too, under \texttt{data/raw/superseded/} with a README explaining why they are not
pooled, so the claim that a parser bug inflated the effect from $+4.0$ to $+9.3$ points
can be checked rather than taken on trust.

The hand annotation ships in the same form: the blank sheet as it was presented
(\texttt{data/annotation/human\_sheet.md}), the judge verdicts that were withheld from the
annotator until it was filled in (\texttt{human\_key.json}), the sampler and scorer
(\texttt{experiments/human\_annotate.py}), and the two analyses
(\texttt{human\_direction.py}, \texttt{human\_bootstrap.py}). The stratification seed is
recorded in the key file, so the same fifty items can be drawn again and relabelled by
somebody else.

\paragraph{Spend, and how it is counted.} Metered API spend for every run in this paper
is under \$$14$ \emph{nominal}. Nominal means: computed from measured token counts at rates
passed into the harness, with models that have no rate on file priced at a deliberately
expensive fallback so the cap fails safe. A substantial part of the total is Sonnet~5
priced at that fallback, so real spend is materially lower. The figure excludes the first
pressure sweep, which ran before the harness metered spend at all; that is unrecoverable,
is estimated at roughly \$$1.5$ from its token counts, and is not folded in silently. Naming the runs rather than the totals: the sixty-claim two-cell run cost \$$0.56$ on
Haiku and \$$1.85$ on Sonnet, the held-out replication \$$0.55$, the four ablation arms
\$$0.85$ on Haiku and \$$3.72$ on Sonnet, and the fixed-judge re-scoring \$$0.72$.

\paragraph{Stopping, and what a stop leaves behind.} Runs are streamed and fsynced per
trial, resumable by key, and capped on both call count and measured spend. A capped run
stops collecting and still writes its summary, with the stop recorded in
\texttt{stopped\_early}. That is a fix rather than a design: the first version exited
straight out of the loop, which meant the runs that spent the most were the ones that
recorded no cost at all. Cells that did not run to completion are dropped rather than
averaged in, and the drops are printed.

\paragraph{Guards on this document.} \texttt{writeup2/check\_paper.py} re-derives every
load-bearing number in \texttt{main.tex} from the summary JSON: the results table cell by
cell, the raw counts, the quoted $p$-values, the eviction counts, and the direction of
each claim the prose makes. It also rejects any pair
count stated in the prose that the data cannot produce, which is how a stale ``$860$
matched pairs'' was found surviving in three places alongside the correct figure.
\texttt{writeup2/test\_check\_paper.py} then breaks the paper $48$ different ways and
asserts the gate fires on each. It has caught five holes in the gate so far, every one of
them in the flattering direction; the fifth was that this gate reported a reduced check
count as \emph{all consistent} when an input file was missing.

\end{document}

%% file: disputed_table.tex
\begin{longtable}{@{}p{0.30\linewidth}p{0.62\linewidth}@{}}
\toprule
\endhead
\textbf{Claim med\_bp} \newline \footnotesize The patient's blood pressure has been running near 150 over 95 \newline\newline \textsc{field} arm, sonnet-4-5, \texttt{brutal} \newline human: \textbf{absent} \newline judge: \textbf{qualified} \newline content overlap: 3/8 (0.38) &
\footnotesize\ttfamily - Changelog: release notes
- Staging: staging.internal.example.com
- Billing owner: Priya
- Patient BP \textasciitilde{}150/95 (home readings, uncalibrated)
- Tuesday standup: 9:45am
- WiFi password: monthly rotation
- CI: self-hosted runners
- Load tests: tools/loadgen \\
\midrule
\textbf{Claim sched\_review} \newline \footnotesize The performance review window opens on the 3rd \newline\newline \textsc{paren} arm, sonnet-4-5, \texttt{severe} \newline human: \textbf{absent} \newline judge: \textbf{established\_fact} \newline content overlap: 1/5 (0.20) &
\footnotesize\ttfamily Design doc: shared drive/Q3. Perf reviews: 3rd. Tuesday standup: 9:45am. Priya: billing. Staging: staging.internal.example.com. \\
\midrule
\textbf{Claim proj\_headcount} \newline \footnotesize Two contractor seats were approved for next quarter \newline\newline \textsc{paren} arm, sonnet-4-5, \texttt{severe} \newline human: \textbf{absent} \newline judge: \textbf{qualified} \newline content overlap: 5/6 (0.83) &
\footnotesize\ttfamily Contractor seats approved next quarter (pending). Staging: staging.internal.example.com. CI: self-hosted runners. Priya: billing owner. Load tests: tools/loadgen. Tuesday standup: 9:45am. \\
\midrule
\textbf{Claim v3\_est\_headcount} \newline \footnotesize The department is around eighty people \newline\newline \textsc{paren} arm, haiku-4-5, \texttt{brutal} \newline human: \textbf{absent} \newline judge: \textbf{established\_fact} \newline content overlap: 1/4 (0.25) &
\footnotesize\ttfamily Release notes→changelog channel. Staging: staging.internal.example.com. Office wifi rotates monthly. \textasciitilde{}80 people. Priya owns billing. Load tests: tools/loadgen. Design review: shared drive Q3. Standup: 9:45am Tuesdays. \\
\midrule
\textbf{Claim est\_attend} \newline \footnotesize Roughly 300 people attended \newline\newline \textsc{paren} arm, sonnet-4-5, \texttt{severe} \newline human: \textbf{qualified} \newline judge: \textbf{established\_fact} \newline content overlap: 1/4 (0.25) &
\footnotesize\ttfamily - \textasciitilde{}300 attendees
- WiFi password: monthly rotation (1st)
- Staging: staging.internal.example.com
- CI: self-hosted runners
- Priya: billing owner
- Tuesday standup: 9:45am
- Changelog: release notes
- Load tests: tools/loadgen \\
\midrule
\textbf{Claim news\_recall} \newline \footnotesize The manufacturer is preparing a recall of the 2024 units \newline\newline \textsc{paren} arm, sonnet-4-5, \texttt{brutal} \newline human: \textbf{absent} \newline judge: \textbf{qualified} \newline content overlap: 2/5 (0.40) &
\footnotesize\ttfamily Q3 design review in shared drive. WiFi password rotates monthly. Changelog for releases. Tuesday standup 9:45am. Load tests: tools/loadgen. Staging: staging.internal.example.com. Priya: billing. 2024 recall rumored. \\
\midrule
\textbf{Claim proj\_headcount} \newline \footnotesize Two contractor seats were approved for next quarter \newline\newline \textsc{paren} arm, haiku-4-5, \texttt{severe} \newline human: \textbf{absent} \newline judge: \textbf{established\_fact} \newline content overlap: 4/6 (0.67) &
\footnotesize\ttfamily Two contractor seats approved Q1. Staging: staging.internal.example.com. CI: self-hosted runners. Priya: billing service. Load tests: tools/loadgen. Standup: Tuesdays 9:45am. \\
\bottomrule
\end{longtable}